# Neural State Classification for Hybrid Systems[*]

Dung Phan[1], Nicola Paoletti[1], Timothy Zhang[1], Radu Grosu[2],
Scott A. Smolka[1], and Scott D. Stoller[1]

[1] Department of Computer Science, Stony Brook University, Stony Brook, NY, USA
[2] Department of Computer Engineering, Technische Universitat Wien, Vienna, Austria

**Abstract.** We introduce the *State Classification Problem* (SCP) for hybrid systems, and present *Neural State Classification* (NSC) as an efficient solution technique. SCP generalizes the model checking problem as it entails classifying each state $s$ of a hybrid automaton as either positive or negative, depending on whether or not $s$ satisfies a given time-bounded reachability specification. This is an interesting problem in its own right, which NSC solves using machine-learning techniques, Deep Neural Networks in particular. State classifiers produced by NSC tend to be very efficient (run in constant time and space), but may be subject to classification errors. To quantify and mitigate such errors, our approach comprises: i) techniques for certifying, with statistical guarantees, that an NSC classifier meets given accuracy levels; ii) tuning techniques, including a novel technique based on *adversarial sampling*, that can virtually eliminate false negatives (positive states classified as negative), thereby making the classifier more conservative. We have applied NSC to six nonlinear hybrid system benchmarks, achieving an accuracy of 99.25% to 99.98%, and a false-negative rate of 0.0033 to 0, which we further reduced to 0.0015 to 0 after tuning the classifier. We believe that this level of accuracy is acceptable in many practical applications, and that these results demonstrate the promise of the NSC approach.

## 1 Introduction

Model checking of hybrid systems is usually expressed in terms of the following reachability problem for hybrid automata (HA): given an HA $\mathcal{M}$, a set of initial states $I$, and a set of unsafe states $U$, determine whether there exists a trajectory of $\mathcal{M}$ starting in an initial state and ending in an unsafe state. The time-bounded version of this problem considers trajectories that are within a given time bound $T$. It has been shown that reachability problems and time-bounded reachability problems for HA are undecidable [16], except for some fairly restrictive classes of HA [7,16]. HA model checkers cope with this undecidability by providing approximate answers to reachability [13].

This paper introduces the *State Classification Problem* (SCP), a generalization of the model checking problem for hybrid systems. Let $\mathbb{B} = \{0,1\}$ be the set of Boolean values. Given an HA $\mathcal{M}$ with state space $S$, time bound $T$, and set of unsafe states $U \subset S$, the SCP problem is to find a function $F^* : S \to \mathbb{B}$ such that for all $s \in S$, $F^*(s) = 1$ if it is possible for $\mathcal{M}$, starting in $s$, to reach a state in $U$ within time $T$;

[*] This material is based on work supported in part by AFOSR Grant FA9550-14-1-0261, NSF Grants CPS-1446832, IIS-1447549, CNS-1445770, CNS-1421893, and CCF-1414078, FWF-NFN RiSE Award, and ONR Grant N00014-15-1-2208.

$F^*(s) = 0$ otherwise. A state $s \in S$ is called *positive* if $F^*(s) = 1$. Otherwise, $s$ is *negative*. We call such a function a *state classifier*.

SCP generalizes the model checking problem. Model checking, in the context of SCP, is simply the problem of determining whether there exists a positive state in the set of initial states. Its intent is not to classify all states in $S$.

Classifying the states of a complex system is an interesting problem in its own right. State classification is also useful in at least two other contexts. First, due to random disturbances, a hybrid system may restart in a random state outside the initial region, and we may wish to check the system's safety from that state. Secondly, a classifier can be used for *online model checking* [26], where in the process of monitoring a system's behavior, one would like to determine, in real-time, the fate of the system going forward from the current (non-initial) state.

This paper shows how deep neural networks (DNNs) can be used for state classification, an approach we refer to as *Neural State Classification* (NSC). An NSC classifier is subject to *false positives* (FPs) – a state $s$ is deemed positive when it is actually negative, and, more importantly, *false negatives* (FNs) – $s$ is deemed negative when it is actually positive.

A well-trained NSC classifier offers high accuracy, runs in constant time (approximately 1 millisecond, in our experiments), and takes constant space (e.g., a DNN with $l$ hidden layers and $n$ neurons only requires functions of dimension $l \cdot n$ for its encoding). This makes NSC classifiers very appealing for applications such as online model checking, a type of analysis subject to strict time and space constraints. NSC classifiers can also be used in runtime verification applications where a low probability of FNs is acceptable, e.g., performance-related system monitoring.

Our approach can also classify states of *parametric* HA by simply encoding each parameter as an additional input to the classifier. This makes NSC more powerful than state-of-the-art hybrid system reachability tools that have little or no support for parametric analysis [12,13]. In particular, we can train a classifier that classifies states of any instance of the parameterized HA, even instances with parameter values not seen during training.

NSC-based classification can be lifted from states to (convex) sets of states by applying output-range estimation [29]. Such techniques can be used to compute safe bounds for the given state region.

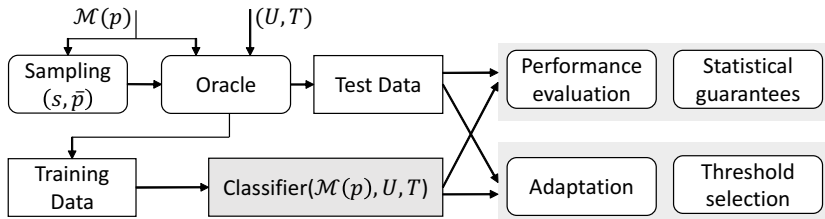

Fig. 1: Overview of the NSC approach.

The NSC method is summarized in Fig. 1. We train the state classifier using supervised learning, where the training examples are derived by sampling the state and parameter spaces according to some distribution. Reachability values for the examples are computed by invoking an oracle, i.e., an hybrid system model checker [13] or a simulator when the system is deterministic.

We consider three sampling strategies: *uniform*, where every state is equi-probable, *balanced*, which seeks to improve accuracy by drawing a balanced number of positive and negative states, and *dynamics-aware*, which assigns to each state the estimated probability that the state is visited in any time-bounded evolution of the system. The choice of sampling strategy depends on the intended application of NSC. For example, in the case of online model checking, dynamics-aware sampling may be the most appropriate. For balanced sampling, we introduce a method to generate arbitrarily large sets of positive samples based on constructing and simulating reverse HAs.

NSC is not limited to DNN-based classifiers. We demonstrate that other machine-learning models for classification, such as support vector machines (SVMs) and binary decision trees (BDTs), also provide powerful solution techniques.

Given the impossibility of training machine-learning models with guaranteed accuracy w.r.t. the true input distribution, we evaluate a trained state classifier by estimating its accuracy, false-positive rate, and false-negative rate (together with their confidence intervals) on a test dataset of fresh samples. This allows us to quantify how well the classifier extrapolates to unseen states, i.e., the probability that it correctly predicts reachability for any state.

Inspired by statistical model checking [24], we also provide statistical guarantees through sequential hypothesis testing to certify (up to some confidence level) that the classifier meets prescribed accuracy levels on unseen data. Note that the systems we consider are *nonprobabilistic*. The statistical guarantees we provide are for the probability that the classifier makes the correct prediction. In contrast, the aim of *probabilistic model checking* [22] and *statistical model checking* [24] is to compute the probability that a probabilistic system satisfies a given correctness property. Relatedly, the focus of neural network (NN) verification [18,11,21] is on proving properties of an NN's output rather than the NN's accuracy.

We also consider two tuning methods that can reduce and virtually eliminate false negatives: a new method called *falsification-guided adaptation* that iteratively re-trains the classifier with false negatives found through adversarial sampling; and *threshold selection*, which adjusts the NN's classification threshold to favor FPs over FNs.

Our experimental results demonstrate the feasibility and promise of our approach, evaluated on six nonlinear hybrid system benchmarks. We consider shallow (1 hidden layer) and deep (3 hidden layers) NNs with sigmoid and ReLU activation functions. Our techniques achieve a prediction accuracy of 99.25% to 99.98% and a false-negative rate of 0.0033 to 0, taking into account the best classifier for each of the six models, with DNNs yielding superior accuracy than shallow NNs, SVMs, and BDTs. We believe that such a range for the FN rate is acceptable in many practical applications, and we show how this can be further improved through tuning of the classifiers.

In summary, our main contributions are the following:

- We introduce the State Classification Problem for hybrid systems.

- We develop the Neural State Classification method for solving the SCP, including techniques for sampling, establishing statistical guarantees on a classifier's accuracy, and reducing its FN rate.
- We introduce a new technique for constructing the *reverse HA* of a given HA, for a general class of HAs, and use reverse HAs to generate balanced training datasets.
- We introduce a *falsification-guided adaptation algorithm* for eliminating FNs, thereby producing conservative state classifiers.
- We provide an extensive evaluation on six nonlinear hybrid system models.

## 2  Problem Formulation

We introduce the problem of learning a state classifier for a hybrid automaton and a bounded reachability property. First, we define these terms.

**Definition 1 (Hybrid automaton).** *A* hybrid automaton *(HA) is a tuple* $\mathcal{M} = (Loc, Var, Init, Flow, Trans, Inv)$, *where* $Loc$ *is a finite set of discrete* locations *(or modes);* $Var = \{x_1, \ldots, x_n\}$ *is a set of continuous* variables, *evaluated over a continuous domain* $X \subseteq \mathbb{R}^n$; $Init \subseteq S(\mathcal{M})$ *is the set of* initial states, *where* $S(\mathcal{M}) = Loc \times X$ *is the* state space *of* $\mathcal{M}$; $Flow : Loc \to (X \to X)$ *is the* flow *function, defining the continuous dynamics at each location;* $Trans$ *is the* transition relation, *consisting of tuples of the form* $(l, g, v, l')$, *where* $l, l' \in Loc$ *are source and target locations, respectively,* $g \subseteq X$ *is the* guard, *and* $v : X \to X$ *is the* reset; $Inv : Loc \to 2^X$ *is the* invariant *at each location.*

We also consider *parameterized* HA in which the flow, guard, reset and invariant may have parameters whose values are constant throughout an execution. We treat parameters as continuous variables with flow equal to zero and identity reset map.

The behavior of an HA $\mathcal{M}$ can be described in terms of its trajectories. A trajectory may start from any state; it does not need to start from an initial state. For time bound $T \in \mathbb{R}^{\geq 0}$, let $\mathbb{T} = [0, T] \subseteq \mathbb{R}^{\geq 0}$ be the time domain.

**Definition 2 (Trajectory [3]).** *For HA* $\mathcal{M} = (Loc, Var, Init, Flow, Trans, Inv)$, *time domain* $\mathbb{T} = [0, T]$, *let* $\rho : \mathbb{T} \to S(\mathcal{M})$ *be a function mapping time instants into states of* $\mathcal{M}$. *For* $t \in \mathbb{T}$, *let* $\rho(t) = (l(t), \mathbf{x}(t))$ *be the state at time* $t$, *with* $l(t)$ *being the location and* $\mathbf{x}(t)$ *the vector of continuous variables. Let* $(\xi_i)_{i=0,\ldots,k} \in \mathbb{T}^{k+1}$ *be the ordered sequence of time points where mode jumps happen, i.e., such that* $\xi_0 = 0$, $\xi_k = T$, *and for all* $i = 0, \ldots, k-1$ *and for all* $t \in [\xi_i, \xi_{i+1})$, $l(t) = l(\xi_i)$. *Then,* $\rho$ *is a* trajectory *of* $\mathcal{M}$ *if it is consistent with the invariants:* $\forall t \in \mathbb{T}.\ \mathbf{x}(t) \in Inv(l(t))$; *flows:* $\forall t \in \mathbb{T}.\ \dot{\mathbf{x}}(t) = Flow(l(t))(\mathbf{x}(t))$; *and transition relation:* $\forall i < k.\ \exists (l(\xi_i), g, v, l(\xi_{i+1})) \in Trans.\ \mathbf{x}(\xi_{i+1}^-) \in g \wedge \mathbf{x}(\xi_{i+1}) = v(\mathbf{x}(\xi_{i+1}^-))$.

**Definition 3 (Time-bounded reachability).** *Given an HA* $\mathcal{M}$, *set of states* $U \subseteq S(\mathcal{M})$, *state* $s \in S(\mathcal{M})$, *and time bound* $T$, *decide whether there exists a trajectory* $\rho$ *of* $\mathcal{M}$ *starting from* $s$ *and* $t \in [0, T]$ *such that* $\rho(t) \in U$, *denoted* $\mathcal{M} \models \mathsf{Reach}(U, s, T)$.

**Definition 4 (Positive and negative states).** *Given an HA* $\mathcal{M}$, *set of states* $U \subseteq S(\mathcal{M})$, *called* unsafe states, *and time bound* $T$, *a state* $s \in S(\mathcal{M})$ *is called* positive *if* $\mathcal{M} \models \mathsf{Reach}(U, s, T)$, *i.e., an unsafe state is reachable from* $s$ *within time* $T$. *Otherwise,* $s$ *is called* negative.

We will use the term *positive (negative) region* for $\mathcal{M}$'s set of positive (negative) states.

**Definition 5 (State classification problem).** *Given an HA $\mathcal{M}$, set of states $U \subseteq S(\mathcal{M})$, and time bound $T$, find a function $F^* : S(\mathcal{M}) \to \mathbb{B}$ such that $F^*(s) = \mathcal{M} \models \mathsf{Reach}(U, s, T)$ for all $s \in S(\mathcal{M})$.*

It is easy to see that the model checking problem for hybrid systems can be expressed as an SCP in which the domain of $F^*$ is the set of initial states $Init$, instead of the whole state space. SCP is therefore a generalization of the model checking problem.

Sample sets are used by NSC to learn state classifiers and to evaluate their performance. Unsafe states are trivially positive (for any $T$), so we exclude them from the sampling domain. Each sample consists of a state $s$ and Boolean $b$ which is the answer to the reachability problem starting from state $s$. We call $(s, 1)$ a *positive sample* and $(s, 0)$ a *negative sample*. Both kinds of samples are generally needed for adequately learning a classifier.

**Definition 6 (Sample set).** *For model $\mathcal{M}$, set of states $U \subseteq S(\mathcal{M})$, and time bound $T$, a* sample set *is any finite set $\{(s, b) \in (S(\mathcal{M}) \setminus U) \times \mathbb{B} \mid b = (\mathcal{M} \models \mathsf{Reach}(U, s, T))\}$*

The derivation of an NSC classifier reduces to a *supervised learning problem*, specifically, a binary classification problem. Given a sample set $D$ called the *training set*, NSC approximates the exact state classifier $F^*$ in Definition 5 by learning a total function $F : (S(\mathcal{M}) \setminus U) \to \mathbb{B}$ from $D$.

Learning $F$ typically corresponds to finding values of $F$'s parameters that minimize some measure of discrepancy between $F$ and the training set $D$. We do not require that the learned function agree with the $D$ on every state that appears in $D$, because this can lead to over-fitting to $D$ and hence poor generalization to other states.

To evaluate the performance of $F$, we use three standard metrics: *accuracy* $P_\mathsf{A}$, i.e, the probability that $F$ produces the correct prediction; the probability of *false positives*, $P_\mathsf{FP}$; and the probability of *false negatives*, $P_\mathsf{FN}$. In safety-critical applications, achieving a low FN rate is typically more important than achieving a low FP rate. Precisely computing these probabilities is, in general, infeasible. We therefore compute an empirical accuracy measure, false-positive rate, and false-negative rate over a *test set* $\mathcal{D}'$ containing $n$ fresh samples not appearing in the training set as follows:

$$\hat{P}_\mathsf{A} = \frac{1}{n} \sum_{(s,b) \in \mathcal{D}'} \mathbf{1}_{F(s)=b} \, , \; \hat{P}_\mathsf{FP} = \frac{1}{n} \sum_{(s,b) \in \mathcal{D}'} \mathbf{1}_{F(s) \wedge \neg b} \, , \; \hat{P}_\mathsf{FN} = \frac{1}{n} \sum_{(s,b) \in \mathcal{D}'} \mathbf{1}_{\neg F(s) \wedge b} \quad (1)$$

where $\mathbf{1}$ is the indicator function. We obtain statistically sound bounds for these probabilities through the Clopper-Pearson method for deriving precise confidence intervals.

## 3 Neural State Classification

This section introduces the main components of the NSC approach.

### 3.1 Neural Networks for Classification

NSC uses *feedforward* neural networks, a type of neural network with one-way connections from input to output layers [23]. NSC uses both *shallow* NNs, with one hidden layer and one output layer, and *deep* NNs, with multiple hidden layers. Additional background on NNs is provided in Appendix C.

An NN defines a real-valued function $F(\mathrm{x})$. When using an NN for classification, a *classification threshold* $\theta$ is specified, and an input vector $\mathrm{x}$ is classified as positive if $F(\mathrm{x}) \geq \theta$, and as negative otherwise.

The theoretical justification for using NNs to solve the SCP is the following. In [17], it is shown that shallow feedforward NNs are universal approximators; i.e., with appropriate parameters, they can approximate any Borel-measurable function arbitrarily well with a finite number of neurons (and just one hidden layer). Under mild assumptions, this also applies to the true state classifier $F^*$ of the SCP (Definition 5). A proof of this claim is given in Appendix A.1 . Arbitrarily high precision might not be achievable in practice, as it would require significantly large training sets and numbers of neurons, and a precise learning algorithm. Nevertheless, NNs are extremely powerful.

### 3.2 Oracles

Given a state (sample) $s$ of an HA $M$, an NSC *oracle* is a procedure for labeling $s$; i.e., for deciding whether $\mathcal{M} \models \mathsf{Reach}(U, s, T)$. NSC utilizes the following oracles.

*Reachability checker.* For nonlinear HA, NSC uses dReal [13], an SMT solver that supports bounded model checking of such HA. dReal provides sound unsatisfiability proofs, but satisfiability is approximated up to a user-defined precision ($\delta$-satisfiability). The oracle first attempts to verify that $s$ is negative by checking $\mathcal{M} \models \mathsf{Reach}(U, s, T)$ for unsatisfiability. If this instance is instead $\delta$-sat, the oracle attempts to prove the unsatisfiability of $\mathcal{M} \models \neg\mathsf{Reach}(U, s, T)$, which would imply that $s$ is positive. The latter instance can also be $\delta$-sat, meaning that this oracle cannot make a decision about $s$. This situation never occurred in our evaluation and can be made less likely by choosing a small $\delta$. If it did occur, our plan would be to conservatively mark the state as positive. The oracle requires an upper bound on the number of discrete jumps to be considered. It supports HAs with Lipschitz continuous dynamics and hyperrectangular continuous domains (i.e., defined as the product of closed intervals), and allows trigonometric and other non-polynomial functions in the initial conditions, guards, invariants, and resets.

*Simulator.* For deterministic systems, we implemented a simulator based on MATLAB's `ode45` variable-step ODE solver. To check reachability, we employ the solver's event-detection method to catch relevant zero-crossing events (i.e., reaching $U$).

*Backwards simulator.* The backwards simulator is not an oracle per se, but, as described in Section 3.3, is central to one of our sampling methods. We first construct the reverse HA according to Definition 7, which is more general than the one for rectangular HAs given in [16]. We use dot-notation to indicate members of a tuple, and lift resets $v$ to sets of states; i.e., $v(X') = \{v(\mathrm{x}) \mid \mathrm{x} \in X'\}$.

**Definition 7 (Reverse HA).** *Given an HA $\mathcal{M}$, its* reverse HA $\overleftarrow{\mathcal{M}}$ *is an HA such that the modes, continuous variables, and invariants are the same as for $\mathcal{M}$, the flows are* *reversed, i.e.,* $\forall (l, \mathrm{x}) \in S(\mathcal{M})$, $\overleftarrow{\mathcal{M}}.Flow(l)(\mathrm{x}) = -\mathcal{M}.Flow(l)(\mathrm{x})$, *and for each transition* $(l, g, v, l') \in \mathcal{M}.Trans$, *the corresponding transition* $(l', \overleftarrow{g}, \overleftarrow{v}, l) \in \overleftarrow{\mathcal{M}}.Trans$ *must be such that* $\overleftarrow{g} = v(g)$ *and* $\overleftarrow{v}$ *is the inverse of $v$ if $v$ is injective; otherwise,* $\overleftarrow{v}$ *updates the continuous state* $\mathrm{x}$ *to any value in the set* $\overleftarrow{v}(\mathrm{x}) = \{\mathrm{x}' \mid \mathrm{x}' \in g \wedge v(\mathrm{x}') = \mathrm{x}\}$.[3]

Although every HA admits a reverse counterpart according to Definition 7, it is clearly impractical to find a reverse reset function $\overleftarrow{v}(\mathrm{x})$ if $v$ is a one-way function. For an example of reversible HA with non-injective reset functions, see the HA and reverse HA in Appendix D.4.

Note that a deterministic HA may admit a nondeterministic reverse HA. Since we classify all states in the state space, we assume that $\mathcal{M}$ and $\overleftarrow{\mathcal{M}}$ can be initialized to any state. We next define the notion of a reverse trajectory $\overleftarrow{\rho}$, which intuitively is obtained by running $\rho$ backwards, starting from $\rho$'s last state and ending with its first state.

**Definition 8 (Reverse trajectory).** *For HA $\mathcal{M}$, time domain $\mathbb{T} = [0, T]$, trajectory $\rho$ with its corresponding sequence of switching time points $(\xi_i)_{i=0,\ldots,k} \in \mathbb{T}^{k+1}$, the reverse trajectory $\overleftarrow{\rho} = (l(t), \mathbf{x}(t))$ of $\rho$ and its corresponding sequence of switching time points $\left(\overleftarrow{\xi}_i\right)_{i=0,\ldots,k} \in \mathbb{T}^{k+1}$ are such that for $i = 0, \ldots, k$, $\overleftarrow{\xi}_i = T - \xi_{k-i}$, and $\forall i < k$, $\overleftarrow{\rho}.l(\overleftarrow{\xi}_i) = \rho.l(\xi_{k-i-1}) \wedge \forall t \in [\overleftarrow{\xi}_i, \overleftarrow{\xi}_{i+1})$, $\overleftarrow{\rho}.l(t) = \overleftarrow{\rho}.l(\overleftarrow{\xi}_i) \wedge \overleftarrow{\rho}.\mathbf{x}(t) = \rho.\mathbf{x}(T - t)$.*

**Theorem 1.** *For an HA $\mathcal{M}$ that admits a reverse HA $\overleftarrow{\mathcal{M}}$, every trajectory $\rho$ of $\mathcal{M}$ is reversible, i.e., the reverse trajectory $\overleftarrow{\rho}$ of $\rho$ is a trajectory of $\overleftarrow{\mathcal{M}}$, and every trajectory $\overleftarrow{\rho}$ of $\overleftarrow{\mathcal{M}}$ is forward-feasible, i.e., the reverse trajectory $\rho$ of $\overleftarrow{\rho}$ is a trajectory of $\mathcal{M}$.*

*Proof.* See Appendix A.2.

Given an unsafe state $u \in U$ of an HA $\mathcal{M}$ that admits a reverse HA $\overleftarrow{\mathcal{M}}$, Theorem 1 allows one to find a positive state $s \in S(\mathcal{M}) \setminus U$ from which $u$ can be reached within time $T$. The method works by simulating multiple trajectories of $\overleftarrow{\mathcal{M}}$ starting in $u$ and up to time $T$. In particular, we explore the reverse trajectories from $u$ through an isotropic random walk, i.e., by choosing uniformly at random, at each step of the simulation, the next transition from those available.

### 3.3 Generation of Training Data and Test Data

We present three sampling methods for generation of training data and test data. Let $\bar{X}$ denote the continuous component of $S(\mathcal{M}) \setminus U$, i.e., without the automaton's location. Recall that model parameters, when present, are expressed as (constant) continuous state variables. They can be sampled independently from the other state variables using appropriate distributions, possibly different from those described below.

*Uniform Sampling.* When the union of mode invariants covers $\bar{X}$, the algorithm first uniformly samples a continuous state $x$ from $\bar{X}$ and then samples a mode $m$ whose invariant is consistent with $x$ (i.e, $x \in Inv(m)$). When the union of mode invariants does not cover $\bar{X}$, we first uniformly sample the mode $m$ and then a continuous state $x \in Inv(m)$. For simplicity, we restrict attention to cases where the region to be sampled is rectangular, although we could use algorithms for uniform sampling of convex polytopes [20]. We use the reachability checker or the simulator (for deterministic systems) to label the sampled states.

---
[3] Technically, for $v$ non-injective, $\overleftarrow{v}$ is in general a nondeterministic reset: $\overleftarrow{v} : X \to 2^X$.

*Balanced Sampling.* In systems where the unsafe states $U$ are a small part of the overall state space, a uniform sampling strategy produces imbalanced datasets with insufficient positive samples, causing the learned classifier to have relatively low accuracy. For such systems, we generate balanced datasets with equal numbers of negative and positive samples as follows. Negative samples are obtained by uniformly sampling states from $S(\mathcal{M}) \setminus U$ and invoking the reachability checker on those states. In this case, the oracle only needs to verify that the sampled state is negative, i.e., to check that $\mathcal{M} \models \mathsf{Reach}(U, s, T)$ is unsatisfiable. For deterministic systems, the simulator is used instead. Positive samples are obtained by uniformly sampling unsafe states $u$ from $U$ and invoking the backwards simulator from $u$.

*Dynamics-Aware Sampling.* This technique generates datasets according to a state distribution expected in a deployed system. It does this by estimating the probability that a state is visited in a trajectory starting from the initial region $\mathsf{Init}$ within time $T'$, where $T' > T$. This is accomplished by uniformly sampling states from $\mathsf{Init}$ and performing a random exploration of the trajectories from those states up to time $T'$. The resulting distribution, called *dynamics-aware state distribution*, is estimated from the multiset of states encountered in those trajectories. In our experiments, we estimate a discrete distribution, but other kinds of distributions (e.g., smooth kernel or piecewise-linear) are also supported. The reachability checker or simulator is used to label states sampled from the resulting distribution. This method typically yields highly unbalanced datasets, and thus should not be applied on its own to generate training data.

### 3.4 Statistical Guarantees with Sequential Hypothesis Testing

Given the infeasibility of training machine-learning models with guaranteed accuracy on unseen data[4], we provide statistical guarantees *a posteriori*, i.e., after training. Inspired by statistical approaches to model checking [24], we employ hypothesis testing to certify that our classifiers meet prescribed levels of accuracy, and FN/FP rates.

We provide guarantees of the form $P_\mathsf{A} \geq \theta_\mathsf{A}$ (i.e., the true accuracy value is above $\theta_\mathsf{A}$), $P_\mathsf{FN} \leq \theta_\mathsf{FN}$ and $P_\mathsf{FP} \leq \theta_\mathsf{FP}$ (i.e., the true rate of FNs and FPs are below $\theta_\mathsf{FN}$ and $\theta_\mathsf{FP}$, respectively). Being based on hypothesis testing, such guarantees are precise up to arbitrary error bounds $\alpha, \beta \in (0, 1)$, such that the probability of Type-I errors (i.e., of accepting $P_x < \theta_x$ when $P_x \geq \theta_x$, where $x \in \{\mathsf{A}, \mathsf{FN}, \mathsf{FP}\}$) is bounded by $\alpha$, and the probability of Type-II errors (i.e., of accepting $P_x \geq \theta_x$ when $P_x < \theta_x$) is bounded by $\beta$. The pair $(\alpha, \beta)$ is known as the *strength* of the test.

To ensure both error bounds simultaneously, the original test $P_x \geq \theta_x$ vs $P_x < \theta_x$ is relaxed by introducing a small indifference region, i.e., we test the hypothesis $H_0$ :

---

[4] Statistical learning theory [28] provides statistical bounds on the generalization error of learn models, but these bounds are very conservative and thus of little use in practice. We use these bounds, however, in the proof of Theorem 2.

$P_x \geq \theta_x + \delta$ against $H_1 : P_x \leq \theta_x - \delta$ for some $\delta > 0$. We use Wald's sequential probability ratio test (SPRT) to provide the above guarantees. SPRT has the important advantage that it does not require a prescribed number of samples to accept one of the two hypotheses, but the decision is made if the available samples provide sufficient evidence. Details of the SPRT can be found in Appendix B.

Note that in statistical model checking, SPRT is used to verify that a probabilistic system satisfies a given property with probability above/below a given threshold. In contrast, in NSC, SPRT is used to verify that the probability of the classifier producing the correct prediction meets a given threshold.

### 3.5 Reducing the False Negative Rate

We discuss strategies to reduce the rate of FNs, the most serious errors from a safety-critical perspective. *Threshold selection* is a simple, yet effective method, which is based on tuning the classification threshold $\theta$ of the NN classifier (see Section 3.1). Decreasing $\theta$ reduces the number of FNs but may increase the number of FPs and thereby reduce overall accuracy. We evaluate the trade-off between accuracy and FNs in Section 4.2.

Another way to reduce the FN rate is to re-train the classifier with unseen FN samples found in the test stage. For this purpose, we devised a whitebox *falsification-guided adaptation algorithm* that, at each iteration, systematically searches for FNs using *adversarial sampling*; i.e., by solving an optimization problem that seeks to maximize the disagreement between predicted and true reachability values. The optimization problem exploits the knowledge it possesses of the function computed by the NN classifier (whitebox approach). FNs found in this way are used to retrain the classifier. The algorithm iterates until the falsifier cannot find any more FNs.

This approach can be viewed as the dual of counterexample-guided abstraction refinement [9]. CEGAR starts from an abstract model that represents an over-approximation of the system dynamics, and uses counterexamples (FPs) to refine the model, thereby reducing the FP rate. Our approach starts from an under-approximation of the positive region (i.e., the set of states leading to a violation) and uses counterexamples (FNs) to make this region more conservative, reducing the FN rate.

We show that under some assumptions about the performance of the classifier and the falsifier, our algorithm converges to an empty set of FNs. Although it may be difficult in practice to guarantee that these assumptions are satisfied, we also show in Section 4.2 that our algorithm performs reasonably well in practice.

For a state $s$, let $F(s) \in [0, 1]$ and $b(s) \in \{0, 1\}$ be the NN prediction and true reachability value, respectively. Let $FN_k$ denote the true set of false negatives (i.e., all states $s$ such that $b(s) = 1$ and $F(s) < \theta$) at the $k$-th iteration of the adaptation algorithm, and let $\hat{FN}_k$ denote the finite subset of $FN_k$ found by the falsifier. The cumulative set of training samples at the $k$-th iteration of the algorithm is denoted $D_k = D \cup \bigcup_{i=1}^{k} \hat{FN}_k$, where $D$ is the set of samples for the initial training of the classifier.

**Assumption 1** *At each iteration $k$, the classifier correctly predicts positive training samples, i.e., $\forall s \in D_k.\ b(s) = 1 \implies F(s) \geq \theta$, and is such that the FP rate w.r.t. training samples is no larger than the FP rate w.r.t. unseen samples.*

**Assumption 2** *At each iteration $k$, the falsifier can always find an FN when it exists, i.e., $FN_k \neq \emptyset \iff \hat{FN}_k \neq \emptyset$.*

**Theorem 2.** *Under Assumptions 1–2, the adaptation algorithm converges to an empty set of FNs with high probability, i.e., for all $\eta \in (0, 1)$, $\Pr(\lim_{k \to \infty} FN_k = \emptyset) \geq 1 - \eta$.*

*Proof. See Appendix A.3 .*

We developed a falsifier that uses a genetic algorithm (GA) [25], a nonlinear optimization method for finding multiple global (sub-)optima. In our case, we indeed have multiple solutions because FN samples are found at the decision boundaries of the classifier, separating the predicted positive and negative regions. Due to the real-valued state space, each set $FN_k$ is either empty or infinite.

FN states have $F(s) - b(s) < -\theta$, while FPs are such that $F(s) - b(s) \geq \theta$. By maximizing the absolute discrepancy $|F(s) - b(s)|$, we can identify both FNs and FPs, where only the former are kept for retraining. Specifically, the GA minimizes the objective function $o(s) = 1/(8 \cdot (F(s) - b(s))^2)$ which, for default threshold $\theta = 0.5$, gives a proportionally higher penalty to correctly predicted states ($0.5 \leq o(s) \leq \infty$) than wrong predictions ($0.125 \leq o(s) \leq 0.5$). We retrain the network with all FN candidates found by the GA, not just the optima.

## 4 Experimental Evaluation

We evaluated our NSC approach on six hybrid-system case studies: a model of the spiking neuron action potential [8], the classic inverted pendulum on a cart, a quadcopter system [15], a cruise controller [8], a powertrain model [19], and a helicopter model [2]. These case studies represent a broad spectrum of hybrid systems and varying degrees of complexity (deterministic, nondeterministic, nonlinear dynamics including trig functions, 2–29 variables, 1–6 modes, 1–11 transitions). Detailed descriptions of the case studies are given in Appendix D.

For all case studies, NSC neural networks were learned using MATLAB's `train` function, with the Levenberg-Marquardt backpropagation algorithm optimizing the mean square error loss function, and the Nguyen-Widrow initialization method for the NN layers. With this setup, we achieved better performance than more standard approaches such as minimizing binary cross entropy using stochastic gradient methods. Training is very fast, taking 2 to 19 seconds for a training dataset with 20,000 samples.

We evaluated the following types of classifiers: sigmoid DNNs (**DNN-S**) with 3 hidden layers of 10 neurons each, with the Tan-Sigmoid activation function for the hidden layers and the Log-Sigmoid activation function for the output layer; shallow NNs (**SNN**), with the same activation functions as **DNN-S** but with one hidden layer of 20 neurons; ReLU DNNs (**DNN-R**), with 3 hidden layers of 10 neurons each, the rectified linear unit (ReLU) activation function for the hidden layers, and the `softmax` function for the output layer; support vector machines with radial kernel (**SVM**); binary decision trees (**BDT**); and a simple classifier that returns the label of the nearest neighbor in the training set (**NBOR**). We also obtained results for DNN ensembles that combine the predictions of multiple DNNs through majority voting. As expected, ensembles outperformed all of the other classifiers. Due to space limitations, these results are omitted.

We learned the classifiers from relatively small datasets, using training sets of 20K samples and test sets of 10K samples, except where noted otherwise. Larger training sets significantly improved classifier performance for only two of the case studies; see Figure 2. Unless otherwise specified, training and test sets are drawn from the same distribution. The NN architecture (numbers of layers and neurons) was chosen empirically. To avoid overfitting, we did not tune the architecture to optimize the performance for our data. We systematically evaluated other architectures (see Appendix E), but found no alternatives with consistently better performance than our default configuration of 3 layers and 10 neurons. We also experimented with 1D Convolutional Neural Networks (CNNs), but they performed worse than the DNN architectures.

In the following, when clear from the context, we omit the modifier "empirical" when referring to accuracy, FN, and FP rates over a test dataset (as opposed to the true accuracy over the state distribution).

### 4.1 Performance Evaluation

Table 1 shows empirical accuracy and FN rate for all classifiers and case studies, using uniform and balanced sampling. We obtain very high classification accuracy for neuron, pendulum, quadcopter and cruise. For these case studies, DNN-based classifiers registered the best performance, with accuracy values ranging between 99.48 % and 99.98 % and FN rates between 0.24% and 0%. Only a minor performance degradation is observed for the shallow neural network **SNN**, with accuracy in the range 98.89-99.85%.

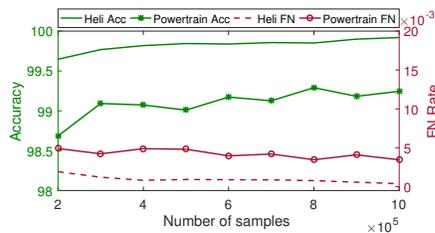

Fig. 2: Performance of DNN-S classifier on helicopter and powertrain models with varying numbers of training samples (uniform sampling).

In contrast, the accuracy for the helicopter and powertrain models is poor if we use only 20K training samples. These models are indeed particularly challenging, owing to their high dimensionality (helicopter) and highly nonlinear dynamics (powertrain). Larger training sets provide considerable improvement in accuracy and FN rate, as shown in Figure 2. For helicopter, accuracy jumps from 98.49% (20K samples) to 99.92% (1M samples), and the FN rate decreases from 0.84% (20K) to 0.04% (1M). For powertrain, accuracy increases from 96.68% (20K) to 99.25% (1M), and the FN rate decreases from 1.28% (20K) to 0.33% (1M).

In general, we found that the NN-based classifiers have superior accuracy compared to support vector machines and binary decision trees. As expected, the nearest-neighbor method demonstrated poor prediction capabilities. No single sampling method provides a clear advantage over the others in terms of accuracy, most likely because training and test sets are drawn from the same distribution.

*Dynamics-aware state distribution.* To evaluate the behavior of the classifiers with the dynamics-aware state distribution (introduced in Section 3.3), we generate training data with a combination of dynamics-aware sampling and either uniform or balanced sampling, because dynamics-aware sampling alone yields unbalanced datasets unsuitable for training. Test data consists exclusively of dynamics-aware samples.

|  | Neuron | | Pendulum | | Quadcopter | | Cruise | | Powertrain | | Helicopter | | |
|---|---|---|---|---|---|---|---|---|---|---|---|---|---|
|  | Acc | FN | Acc | FN | Acc | FN | Acc | FN | Acc | FN | Acc | FN | |
| **DNN-S** | **99.81** | **0.1** | **99.98** | **0** | 99.83 | 0.1 | 99.95 | 0.01 | **96.68** | 1.28 | **98.49** | 0.84 | |
| **DNN-R** | 99.52 | 0.29 | 99.93 | 0.04 | **99.89** | **0.06** | **99.98** | **0** | 96.21 | **1.08** | 98 | 0.96 | Uniform |
| **SNN** | 99.17 | 0.43 | 99.81 | **0** | 99.85 | 0.08 | 99.84 | 0.15 | 96.02 | 1.37 | 97.69 | 1.25 | |
| **SVM** | 98.73 | 0.75 | 99.84 | **0** | 97.33 | 0.69 | 99.88 | 0.1 | 92.26 | 3.48 | 95.58 | 2.42 | |
| **BDT** | 99.3 | 0.37 | 99.6 | 0.17 | 99.52 | 0.2 | 99.84 | 0.08 | 95.59 | 2.19 | 80.07 | 9.8 | |
| **NBOR** | 97.03 | 1.22 | 99.69 | 0.14 | 99.53 | 0.25 | 99.49 | 0.33 | 71.44 | 14.51 | 67.39 | 16.98 | |
|  | Acc | FN | Acc | FN | Acc | FN | Acc | FN | Acc | FN | Acc | FN | |
| **DNN-S** | **99.83** | **0.12** | **99.89** | **0** | **99.82** | 0.04 | 99.94 | **0** | **97.2** | **0.86** | **98.24** | **0.79** | |
| **DNN-R** | 99.48 | 0.24 | 99.63 | 0.01 | 99.67 | 0.09 | **99.95** | **0** | 96.07 | 1.24 | 97.91 | 1.2 | |
| **SNN** | 98.89 | 0.69 | 99.2 | **0** | 99.49 | **0.01** | 99.6 | **0** | 95.21 | 1.79 | 97.58 | 1.16 | Balanced |
| **SVM** | 98.63 | 0.78 | 99.37 | **0** | 96.93 | 0.2 | 99.61 | **0** | 91.84 | 3.3 | 95.36 | 1.85 | |
| **BDT** | 99.07 | 0.45 | 99.46 | 0.05 | 99.36 | 0.22 | 99.9 | 0.03 | 95.86 | 2.4 | 79.03 | 10.26 | |
| **NBOR** | 96.95 | 1.62 | 99.51 | 0.04 | 99.11 | 0.56 | 99.47 | 0.11 | 71.33 | 13.99 | 65.18 | 17.48 | |

Table 1: Empirical accuracy (Acc) and FN rate of the state classifiers for each case study, classifier type, and sampling method. Values are in percentages. For each measure and sampling method, the best result is highlighted in bold. False positives and confidence intervals are reported in Tables 5 and 6 of the Appendix.

Table 2 shows that the classifiers yield accuracy values comparable to those of Table 1 (compiled with balanced and uniform distributions) for all case studies. We see that the powertrain model attains 100% accuracy, indicating that its dynamics-aware distribution favors states that are easy enough for the DNN to classify correctly.

|  | Neuron | Pendulum | Quadcopter | Cruise | Helicopter | Powertrain |
|---|---|---|---|---|---|---|
| **Unif+Dyn-aware** | 99.91 (+0.1) | 99.93 (-0.05) | 99.84 (+0.01) | 99.14 (-0.81) | 98.77 (+0.28) | 100 (+3.32) |
| **Bal+Dyn-aware** | 99.8 (-0.03) | 99.88 (-0.01) | 99.79 (-0.03) | 99.35 (-0.59) | 98.46 (+0.22) | 100 (+2.8) |

Table 2: Empirical accuracy of DNN-S classifiers tested on 10K dynamics-aware samples and trained with 20K samples. Each row corresponds to a different training distribution. **Unif+Dyn-aware** and **Bal+Dyn-aware** were obtained by combining 10K uniform/balanced samples with 10K dynamics-aware samples. In parenthesis is the accuracy difference with the corresponding classifier from Table 1.

*Parametric analysis.* We show that NSC works effectively for parametric systems, being able to classify states in models with parameter values not seen during training. We derive parametric versions of the neuron model by turning constants $a, b, c, d, I$ (see Appendix D.1) into parameters uniformly distributed in the $\pm 50\%$ interval around their default value.

|  | Num. of parameters | | | | |
|---|---|---|---|---|---|
|  | 1 | 2 | 3 | 4 | 5 |
| $\hat{P}_A$ | 99.8 | 99.7 | 97.9 | 98.1 | 97.8 |
| $\hat{P}_{FN}$ | 0.2 | 0.2 | 1.6 | 1.3 | 1.5 |

Table 3: Empirical accuracy ($\hat{P}_A$) and FN rate ($\hat{P}_{FN}$) for DNN-S classifier for neuron model with increasing number of parameters.

Table 3 shows the accuracy and FN rates for DNN-S, trained with 110K samples for models with increasing numbers of parameters, which are increasingly long prefixes of the sequence $a, b, c, d, I$. We achieve very high accuracy ($\geq 99.7\%$) for up to two parameters. For three to five parameters, the accuracy decreases but stays relatively high (around 98%), suggesting that larger training sets are required for these cases.

Indeed the input space grows exponentially in the number of parameters, while we kept the size of the training set constant.

*Statistical guarantees.* We use SPRT (Section 3.4) to provide statistical guarantees for four case studies, each trained with 20K balanced samples. See Table 4. We assess two properties certifying that the *true* (not empirical) accuracy and FNs meet given performance levels: $P_\mathsf{A} \geq 99.7\%$, and $P_\mathsf{FN} \leq 0.2\%$. We omit the helicopter and powertrain models from this assessment, because performance results for these models are clearly outside the desired levels when only 20K samples are used for training.

The only classifier that guarantees these performance levels for all case studies is the sigmoid DNN. We also observe that a small number of samples suffices to obtain statistical guarantees with the given strength: only 3 out of 48 tests needed more than 10K samples to reach a decision.

|        | **Neuron** | | **Pendulum** | | **Quadcopter** | | **Cruise** | |
|--------|:---:|:---:|:---:|:---:|:---:|:---:|:---:|:---:|
|        | $P_\mathsf{A} \geq \theta_\mathsf{A}$ | $P_\mathsf{FN} \leq \theta_\mathsf{FN}$ | $P_\mathsf{A} \geq \theta_\mathsf{A}$ | $P_\mathsf{FN} \leq \theta_\mathsf{FN}$ | $P_\mathsf{A} \geq \theta_\mathsf{A}$ | $P_\mathsf{FN} \leq \theta_\mathsf{FN}$ | $P_\mathsf{A} \geq \theta_\mathsf{A}$ | $P_\mathsf{FN} \leq \theta_\mathsf{FN}$ |
| **DNN-S** | ✓ (5800) | ✓ (2900) | ✓ (2300) | ✓ (2300) | ✓ (4400) | ✓ (2300) | ✓ (3000) | ✓ (2300) |
| **DNN-R** | ✗ (3600) | ✗ (8600) | ✓ (15500) | ✓ (4000) | ✗ (1400) | ✓ (7300) | ✓ (3000) | ✓ (2300) |
| **SNN** | ✗ (700) | ✗ (1000) | ✗ (2900) | ✓ (2300) | ✗ (1500) | ✓ (3400) | ✗ (3600) | ✓ (2300) |
| **SVM** | ✗ (400) | ✗ (600) | ✗ (6600) | ✓ (2300) | ✗ (200) | ✗ (5300) | ✗ (3400) | ✓ (2300) |
| **BDT** | ✗ (1700) | ✗ (3300) | ✗ (6300) | ✓ (15000) | ✗ (800) | ✗ (1100) | ✓ (2700) | ✓ (2900) |
| **NBOR** | ✗ (300) | ✗ (300) | ✗ (28500) | ✓ (2900) | ✗ (1000) | ✗ (1300) | ✗ (3400) | ✗ (2300) |

Table 4: Statistical guarantees based on the SPRT. Samples were generated using balanced sampling. In parenthesis are the number of samples required to reach the decision. Parameters of the test are $\alpha = \beta = 0.01$ and $\delta = 0.001$. Thresholds are $\theta_\mathsf{A} = 99.7\%$ and $\theta_\mathsf{FN} = 0.2\%$.

### 4.2 Reducing the False Negative Rate

*Falsification-guided adaptation.* We evaluate the benefits of adaptation by incrementally adapting the trained NNs with false negative samples (see Section 3.5). At each iteration, we run our GA-based falsifier to find FN samples, which are then used to adapt the DNN. The adaptation loop terminates when the falsifier cannot find a FN.

We employ MATLAB's `adapt` function with gradient descent learning algorithm and learning rates of 0.0005 for neuron and 0.003 for quadcopter, helicopter, and powertrain. For neuron and quadcopter, we use DNN-S classifiers trained with 20K balanced samples. We use DNN-S trained with 1M balanced samples for helicopter, and DNN-S trained with 1M uniform samples for powertrain, because these classifiers have the best accuracy before adaptation. To measure adaptation performances, we test the DNNs on 10K samples after each iteration of adaptation. Fig. 3 shows how accuracy, FNs and FPs of the classifier evolve at each adaptation step. For the neuron, quadcopter, and helicopter case studies, our falsification-guided adaptation algorithm works well to eliminate the FN rate at the cost of a slight increase in the FP rate after only 5-10 iterations. In these case studies, the number of FNs found by the falsifier decreases quickly from hundreds or thousands to zero. For powertrain, the number of FNs found by the falsifier stays almost constant at about 70 on average at each iteration. After 150 iterations, FN rate of the powertrain DNN decreases slowly from 0.33% to 0.15%.

Figure 4 visualizes the effects of adaptation on the **DNN-S** classifier for the neuron case study. Fig. 4 (a) shows the prediction of the DNN after training with 20K samples. Fig. 4 (b) shows the prediction of the DNN after adaptation. We see that adaptation expands the predicted positive region to enclose all previous FN samples, i.e., they are correctly re-classified as positive. The enlarged positive region also means the adapted DNN is more conservative, producing more FPs as shown in Fig. 4 (b).

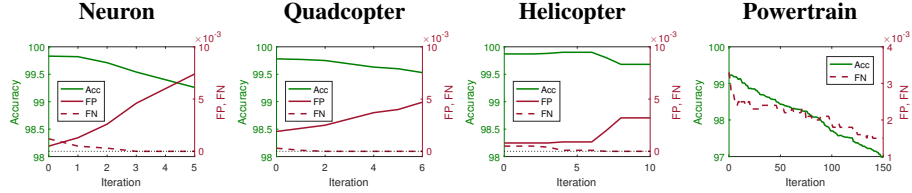

Fig. 3: Impact of incremental adaptation on empirical accuracy, FN and FP rates. FP-rate curve for powertrain is omitted to allow using a scale that shows the decreasing trend of the FN rate. The FP rate for powertrain increases from 0.48% to 2.89%.

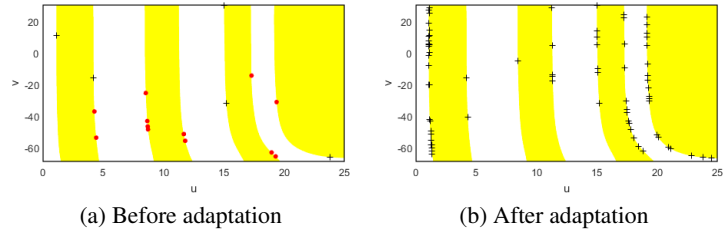

(a) Before adaptation      (b) After adaptation

Fig. 4: Effects of adaptation on the DNN-S for the neuron case study. The white region is the predicted negative region. The yellow region is the predicted positive region. Red dots are FN samples. Crosses are FP samples.

*Threshold selection.* We show that threshold selection can considerably reduce the FN rate. Figure 5 shows the effect of threshold selection on accuracy, FN rate, and FP rate for classifier **DNN-S** trained with uniform sampling (20K samples for neuron and quadcopter, 1M samples for helicopter and powertrain). Pendulum and cruise control case studies are excluded as they have low FN rate ($\leq 0.01\%$) prior to threshold selection.

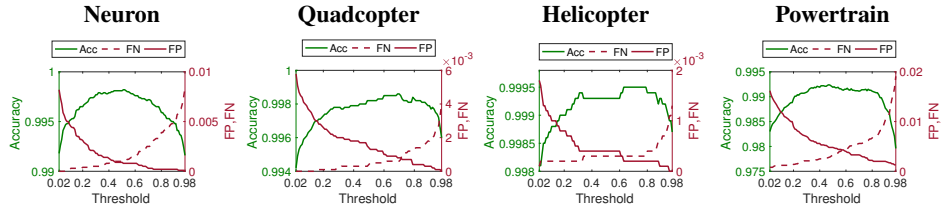

Fig. 5: Impact of classification threshold on empirical accuracy, FN rate, and FP rate.

For the neuron case study, selecting $\theta = 0.32$ reduces the FN rate from $10^{-3}$ to $5 \cdot 10^{-4}$, with an accuracy loss of only $0.02\%$. With $\theta = 0.06$, we obtain a zero FN rate and a minor accuracy loss of $0.37\%$. For quadcopter, selecting $\theta = 0.28$ decreases the FN rate from $4 \cdot 10^{-4}$ to $10^{-4}$, with an accuracy loss of just $0.02\%$. Selecting $\theta = 0.16$ yields zero FN rate and accuracy loss of just $0.12\%$. For helicopter, selecting $\theta = 0.33$ reduces the FN rate from $3 \cdot 10^{-4}$ to $2 \cdot 10^{-4}$, with an *accuracy gain* of $0.01\%$. For powertrain, $\theta = 0.34$ yields a good trade-off between FN rate reduction (from $3.3 \cdot 10^{-3}$ to $2.1 \cdot 10^{-3}$) and accuracy loss ($0.1\%$).

## 5 Related Work

Related work includes techniques for *simulation-based verification*, which enables rigorous system analysis from finitely many executions. Statistical model checking [24] for the verification of probabilistic systems with statistical guarantees is an example of this form of verification. Simulation is also used for falsification and reachability analysis of hybrid systems [1,2]. Our NSC approach also simulates system executions (when the system is deterministic), but for the purpose of learning a state classifier.

Other applications of *machine learning in verification* include parameter synthesis of stochastic systems [5], techniques for inferring temporal logic specifications from examples [4], synthesis of invariants for program verification [27,14], and reachability checking of Markov decision processes [6].

For safety-critical applications, *verification of NNs* has become a very active area, with a focus on the derivation of adversarial inputs (i.e., those that induce incorrect predictions). Most such approaches rely on SMT-based techniques [18,21,11], while sampling-based methods are used in [10] for the analysis of NN components "in the loop" with cyber-physical system models. Similarly, our adaptation method systematically searches for adversarial inputs (FNs) to render the classifier more conservative. A related problem is that of range estimation [29], i.e., computing safe and tight enclosures for the predictions of an NN over a (convex) input region. Such methods could be used to extend NSC classification to sets of states.

## 6 Conclusions

We have introduced the state classification problem for hybrid systems and offered a highly efficient solution based on neural state classification. NSC features high accuracy and low false-negative rates, while including techniques for virtually eliminating such errors and for certifying an NSC classifier's performance with statistical guarantees. Plans for future work include considering more expressive temporal properties and extending our approach to stochastic hybrid systems.

# A Proofs

## A.1 Shallow NNs are universal approximators for state classification

We first need the following mild assumptions:

1. $S(\mathcal{M})$ is compact and Euclidean. This is satisfied in any practical instance of the SCP (after mapping discrete locations into $\mathbb{R}$), e.g., by specifying the state space as a convex set.
2. $\mathbb{T} \subset \mathbb{Q}$, i.e., the time domain is rational (and not dense). This means that trajectories (and thus, reachability) are not considered at irrational time points, an assumption that is reasonable and at the core of any hybrid system simulation/verification tool.
3. The true state classifier $F^*$ of Definition 5 is such that, for any $s \in S(\mathcal{M})$, $F^*(s')$ is constant for all $s' \in \mathcal{B}(s)$, where $\mathcal{B}(s)$ is the smallest rectangle with rational endpoints containing $s$. This is also quite reasonable, as reachability tools typically work by replacing $s$ with the smallest box with endpoints that are representable in machine arithmetic.

*Proof.* To prove that shallow NNs are universal approximators for the true state classifier $F^*$, we need to demonstrate that $F^*$ is defined over a compact set (true by assumption 1), and that $F^*$ is Borel-measurable.

The latter is proved by showing that the positive (or, equivalently, negative) region $F^{*-1}(1)$ is a Borel set, i.e., it is formed from countable union, intersection and complement of closed (or open) sets.

Note that the set of states reachable from $s$ within time $T$ is Borel as it is defined by $R(s, T) = \bigcup_{t \leq T} \{\rho(t) \mid \rho \text{ is a trajectory from } s\}$ where $\rho(t)$ is the state at time $t$ in trajectory $\rho$. By assumption 2, the outermost union is countable and thus, $R(s, T)$ is Borel because, due to the finite number of HA locations, there is a finite number of trajectories from $s$. Then, we can define the positive region by $F^{*-1}(1) = \{s \in S(\mathcal{M}) \mid R(s, T) \cap U \neq \emptyset\}$, which by assumption 3, is equivalent to $F^{*-1}(1) = \{s' \in S(\mathcal{M}) \cap \mathbb{Q}^n \mid R(s', T) \cap U \neq \emptyset\}$. The latter is constructed by countable union (we consider only rational states $s'$), and thus, $F^{*-1}(1)$ is Borel. □

## A.2 Proof of Theorem 1

In this section, we give a proof for the first part of Theorem 1, i.e., every trajectory of $\mathcal{M}$ is reversible. The proof for the second part, i.e., every trajectory of $\overrightarrow{\mathcal{M}}$ is forward-feasible, is similar.

*Proof.* Consider a trajectory $\rho(t) = (l(t), \mathbf{x}(t))$ of $\mathcal{M}$ starting in state $s(0) = (\rho.l(0), \rho.\mathbf{x}(0))$ and ending in state $s(T) = (\rho.l(T), \rho.\mathbf{x}(T))$, and its corresponding sequence of switching time points $(\xi_i^\rho)_{i=0...k}$. Let $\overleftarrow{\rho}(t)$ be the reverse trajectory of $\rho$, and let $(\overleftarrow{\xi}_i)_{i=0...k}$ be the corresponding switching time points of $\overleftarrow{\rho}$. We prove that $\overleftarrow{\rho}$ is a trajectory of $\overleftarrow{\mathcal{M}}$, i.e., we prove that:

1. $\overleftarrow{\xi}_0 = 0$, $\overleftarrow{\xi}_i < \overleftarrow{\xi}_{i+1} \forall i = 0, \ldots, k-1$, and $\overleftarrow{\xi}_k = T$, i.e., the sequence of reverse switching time points increases, starts with 0 and ends with $T$.

2. $\overleftarrow{\rho}.l(0) = \rho.l(T)$, $\overleftarrow{\rho}.\mathbf{x}(0) = \rho.\mathbf{x}(T)$, $\overleftarrow{\rho}.l(T) = \rho.l(0)$, $\overleftarrow{\rho}.\mathbf{x}(T) = \rho.\mathbf{x}(0)$, i.e., the reverse trajectory starts in $\rho$'s terminal state and ends in $\rho$'s initial sate.
3. $\forall i = 0, \ldots, k-1, \forall t \in [\overleftarrow{\xi}_i, \overleftarrow{\xi}_{i+1})$. $\overleftarrow{\rho}.\dot{\mathbf{x}}(t) = \overleftarrow{\mathcal{M}}.Flow(\overleftarrow{\rho}.l(\overleftarrow{\xi}_i))(\overleftarrow{\rho}.\mathbf{x}(t))$, i.e., the continuous evolution is consistent with the differential equation of the corresponding location in $\overleftarrow{\mathcal{M}}$.
4. $\forall i = 0, \ldots, k-1, \forall t \in [\overleftarrow{\xi}_i, \overleftarrow{\xi}_{i+1})$. $\overleftarrow{\rho}.\mathbf{x}(t) \in \overleftarrow{\mathcal{M}}.Inv(\overleftarrow{\rho}.l(\overleftarrow{\xi}_i))$, i.e., $\overleftarrow{\rho}$ is consistent with $\overleftarrow{\mathcal{M}}$'s invariants.
5. $\forall i = 0, \ldots, k-1. \exists (\overleftarrow{\rho}.l(\overleftarrow{\xi}_i), \overleftarrow{g}, \overleftarrow{v}, \overleftarrow{\rho}.l(\overleftarrow{\xi}_{i+1})) \in \overleftarrow{\mathcal{M}}.Trans : \overleftarrow{\rho}.\mathbf{x}(\overleftarrow{\xi}_{i+1}^-) \in \overleftarrow{g} \land \overleftarrow{\rho}.\mathbf{x}(\overleftarrow{\xi}_{i+1}) \in \overleftarrow{v}(\overleftarrow{\rho}.\mathbf{x}(\overleftarrow{\xi}_{i+1}^-))$, where $\overleftarrow{\rho}.\mathbf{x}(\overleftarrow{\xi}_{i+1}^-) = \lim_{t \to \overleftarrow{\xi}_{i+1}^-} \overleftarrow{\rho}.\mathbf{x}(t)$, i.e., $\overleftarrow{\rho}$ is consistent with $\overleftarrow{\mathcal{M}}$'s transition relations. Note that we take liberties with notation here and in the proof that follows: if $\overleftarrow{v}$ is deterministic, then $\overleftarrow{\rho}.\mathbf{x}(\overleftarrow{\xi}_{i+1}) \in \overleftarrow{v}(\overleftarrow{\rho}.\mathbf{x}(\overleftarrow{\xi}_{i+1}^-))$ really means $\overleftarrow{\rho}.\mathbf{x}(\overleftarrow{\xi}_{i+1}) = \overleftarrow{v}(\overleftarrow{\rho}.\mathbf{x}(\overleftarrow{\xi}_{i+1}^-))$.

**Proof of 1.** From the definition of trajectory $\xi_0 = 0$, $\xi_k = T$, and $\xi_i < \xi_{i+1}, \forall i = 0, \ldots, k-1$. Therefore, from the definition of reverse trajectory, $\overleftarrow{\xi}_0 = T - \xi_k = 0$, $\overleftarrow{\xi}_k = T - \xi_0 = T$, and $\overleftarrow{\xi}_i = T - \xi_{k-i} < T - \xi_{k-i-1} = \overleftarrow{\xi}_{i+1}, \forall i = 0, \ldots, k-1$.

**Proof of 2.** From the definition of reverse trajectory, we readily obtain $\overleftarrow{\rho}.l(0) = \rho.l(T)$, $\overleftarrow{\rho}.\mathbf{x}(0) = \rho.\mathbf{x}(T)$, $\overleftarrow{\rho}.l(T) = \rho.l(0)$, $\overleftarrow{\rho}.\mathbf{x}(T) = \rho.\mathbf{x}(0)$.

**Proof of 3.** Consider the time interval $(\overleftarrow{\xi}_i, \overleftarrow{\xi}_{i+1})$ of the reverse trajectory and the corresponding time interval $(\xi_{k-i-1}, \xi_{k-i})$ of the forward trajectory. We have $\forall t \in (\xi_{k-i-1}, \xi_{k-i}), \rho.\dot{\mathbf{x}}(t) = \mathcal{M}.Flow(\rho.l(\xi_{k-i-1}))(\rho.\mathbf{x}(t))$. That means, for all $t \in (\xi_{k-i-1}, \xi_{k-i})$, we have the following equivalent equations:

$$\rho.\mathbf{x}(\xi_{k-i}) = \rho.\mathbf{x}(t) + \int_t^{\xi_{k-i}} \mathcal{M}.Flow(\rho.l(\xi_{k-i-1}))(x(\tau))d\tau$$

$$\rho.\mathbf{x}(t) = \rho.\mathbf{x}(\xi_{k-i}) - \int_t^{\xi_{k-i}} \mathcal{M}.Flow(\rho.l(\xi_{k-i-1}))(x(\tau))d\tau$$

$$\rho.\mathbf{x}(t) = \rho.\mathbf{x}(\xi_{k-i}) - \int_{T-\xi_{k-i}}^{T-t} \mathcal{M}.Flow(\rho.l(\xi_{k-i-1}))(x(\tau))d\tau$$

$$\rho.\mathbf{x}(t) = \rho.\mathbf{x}(\xi_{k-i}) + \int_{T-\xi_{k-i}}^{T-t} -\mathcal{M}.Flow(\rho.l(\xi_{k-i-1}))(x(\tau))d\tau$$

$$\rho.\mathbf{x}(t) = \rho.\mathbf{x}(\xi_{k-i}) + \int_{T-\xi_{k-i}}^{T-t} -\mathcal{M}.Flow(\overleftarrow{\rho}.l(\overleftarrow{\xi}_i))(x(\tau))d\tau$$

$$\overleftarrow{\rho}.\mathbf{x}(T-t) = \overleftarrow{\rho}.\mathbf{x}(\overleftarrow{\xi}_i) + \int_{\overleftarrow{\xi}_i}^{T-t} \overleftarrow{\mathcal{M}}.Flow(\overleftarrow{\rho}.l(\overleftarrow{\xi}_i))(x(\tau))d\tau$$

The last equation means for all $t \in (\overleftarrow{\xi}_i, \overleftarrow{\xi}_{i+1})$, we have:

$$\overleftarrow{\rho}.\dot{\mathbf{x}}(t) = \overleftarrow{\mathcal{M}}.Flow(\overleftarrow{\rho}.l(\overleftarrow{\xi}_i))(\overleftarrow{\rho}.\mathbf{x}(t)). \tag{2}$$

Since $\overleftarrow{\xi}_i$ is a switching time point, $x(\overleftarrow{\xi}_i)$ is the initial condition for the evolution of $x$ in mode $\overleftarrow{\rho}.l(\overleftarrow{\xi}_i)$. Therefore, Eq. 2 holds for all $t \in [\overleftarrow{\xi}_i, \overleftarrow{\xi}_{i+1})$.

**Proof of 4.** Consider the time interval $[\overleftarrow{\xi}_i, \overleftarrow{\xi}_{i+1})$ of the reverse trajectory and the corresponding time interval $[\xi_{k-i-1}, \xi_{k-i})$ of the forward trajectory. We have $\overleftarrow{\rho}.\mathbf{x}(t) = \rho.\mathbf{x}(T-t) \in \mathcal{M}.Inv(\rho.l(\xi_{k-i-1})) = \mathcal{M}.Inv(\overleftarrow{\rho}.l(\overleftarrow{\xi}_i)) = \overleftarrow{\mathcal{M}}.Inv(\overleftarrow{\rho}.l(\overleftarrow{\xi}_i))$

**Proof of 5.** Consider a time point $\overleftarrow{\xi}_{i+1}$, $0 \leq i < k$, when the reverse trajectory jumps from mode $\overleftarrow{\rho}.l(\overleftarrow{\xi}_i)$ to mode $\overleftarrow{\rho}.l(\overleftarrow{\xi}_{i+1})$. According to definition of reverse trajectory, this time point corresponds to the time point $\xi_{k-i-1}$ in the forward trajectory when the transition $(\rho.l(\xi_{k-i-2}), g, v, \rho.l(\xi_{k-i-1}))$ occurs. By definition of reverse HA, there exists a transition $(\rho.l(\xi_{k-i-1}), \overleftarrow{g}, \overleftarrow{v}, \rho.l(\xi_{k-i-2}))$ in $\overleftarrow{\mathcal{M}}.Trans$. By definition of reverse trajectory, this transition is the same as $(\overleftarrow{\rho}.l(\overleftarrow{\xi}_i), \overleftarrow{g}, \overleftarrow{v}, \overleftarrow{\rho}.l(\xi_{i+1}))$. We prove that $\overleftarrow{\rho}.\mathbf{x}(\overleftarrow{\xi}^-_{i+1}) \in \overleftarrow{g}$ and $\overleftarrow{\rho}.\mathbf{x}(\overleftarrow{\xi}_{i+1}) \in \overleftarrow{v}(\overleftarrow{\rho}.\mathbf{x}(\overleftarrow{\xi}^-_{i+1}))$.

We have $\rho.\mathbf{x}(\xi_{k-i-1}) = v(\rho.\mathbf{x}(\xi^-_{k-i-1}))$ and $\overleftarrow{v}$ is the inverse of $v$, therefore:

$$\rho.\mathbf{x}(\xi^-_{k-i-1}) \in \overleftarrow{v}(\rho.\mathbf{x}(\xi_{k-i-1})) \tag{3}$$

Since $\overleftarrow{g} = v(g)$ and $\rho.\mathbf{x}(\xi_{k-i-1}) = v(\rho.\mathbf{x}(\xi^-_{k-i-1}))$, we also have:

$$\rho.\mathbf{x}(\xi_{k-i-1}) \in \overleftarrow{g} \tag{4}$$

Consider the interval $[\xi_{k-i-1}, \xi_{k-i})$, we have the following equivalent equations.

$$\rho.\mathbf{x}(\xi_{k-i-1}) = \lim_{t \to \xi^+_{k-i-1}} \rho.\mathbf{x}(\xi_{k-i}) - \int_t^{\xi_{k-i}} \mathcal{M}.Flow(\rho.l(\xi_{k-i-1}))(\mathbf{x}(\tau))d\tau$$

$$\rho.\mathbf{x}(\xi_{k-i-1}) = \lim_{T-t \to (T-\xi_{k-i-1})^-} \rho.\mathbf{x}(\xi_{k-i}) - \int_{T-\xi_{k-i}}^{T-t} \mathcal{M}.Flow(\rho.l(\xi_{k-i-1}))(\mathbf{x}(\tau))d\tau$$

$$\rho.\mathbf{x}(\xi_{k-i-1}) = \lim_{t \to (\overleftarrow{\xi}_{i+1})^-} \overleftarrow{\rho}.\mathbf{x}(\overleftarrow{\xi}_i) - \int_{\overleftarrow{\xi}_i}^t \mathcal{M}.Flow(\rho.l(\xi_{k-i-1}))(\mathbf{x}(\tau))d\tau$$

$$\rho.\mathbf{x}(\xi_{k-i-1}) = \lim_{t \to (\overleftarrow{\xi}_{i+1})^-} \overleftarrow{\rho}.\mathbf{x}(\overleftarrow{\xi}_i) + \int_{\overleftarrow{\xi}_i}^t \overleftarrow{\mathcal{M}}.Flow(\overleftarrow{\rho}.l(\overleftarrow{\xi}_i))(\mathbf{x}(\tau))d\tau$$

$$\rho.\mathbf{x}(\xi_{k-i-1}) = \overleftarrow{\rho}.\mathbf{x}(\overleftarrow{\xi}^-_{i+1})$$

Combining with Eq. 4, we obtain $\overleftarrow{\rho}.\mathbf{x}(\overleftarrow{\xi}^-_i) \in \overleftarrow{g}$.

Similarly, we can show that $\overleftarrow{\rho}.\mathbf{x}(\overleftarrow{\xi}_{i+1}) = \rho.\mathbf{x}(\xi^-_{k-i-1})$. Combining with Eq. 3, we have $\overleftarrow{\rho}.\mathbf{x}(\overleftarrow{\xi}_{i+1}) \in \overleftarrow{v}(\overleftarrow{\rho}.\mathbf{x}(\overleftarrow{\xi}^-_{i+1}))$.

### A.3 Proof of Theorem 2

*Proof.* The proof follows from Vapnik's bounds for the generalization error of learning models [28], stating that $\Pr(\hat{P}^{\text{te}}_{\text{err}} \leq \hat{P}^{\text{tr}}_{\text{err}} + f(m, D, \eta)) \geq 1 - \eta$, where $\hat{P}^{\text{te}}_{\text{err}}$ is the test error, $\hat{P}^{\text{tr}}_{\text{err}}$ is the training error, $m$ is the number of training samples, and $D$ is the VC dimension which is fixed and characterizes the expressiveness of the learning model. In particular, $\lim_{m \to \infty} f(m, D, \eta) = 0$, as the expression is defined by $f(m, D, \eta) = \sqrt{\frac{1}{m}(D(\log(2m/d) + 1) - \log(\eta/4))}$. Rewrite $\hat{P}^{\text{te}}_{\text{err}} = \hat{P}^{\text{te}}_{\text{FP}} + \hat{P}^{\text{te}}_{\text{FN}}$ as the sum of FP and FN errors, and similarly for $\hat{P}_{\text{err}}^{\text{tr}}$. By assumption 1, $\hat{P}_{\text{FP}}^{\text{tr}} \leq \hat{P}_{\text{FP}}^{\text{te}}$ and $\hat{P}_{\text{FN}}^{\text{tr}} = 0$. Thus, we can rewrite Vapnik's bounds as $\Pr(\hat{P}_{\text{FN}}^{\text{te}} \leq f(m, D, \eta)) \geq 1 - \eta$, implying that $\Pr(\lim_{m \to \infty} \hat{P}_{\text{FN}}^{\text{te}} = 0) \geq 1 - \eta$. By assumption 2, the limit $m \to \infty$ is attainable (the falsifier can always produce FNs if available), unless the falsifier cannot find any, which implies $FN_k = \emptyset$ in absolute terms. Finally note that $\lim_{m \to \infty} \hat{P}_{\text{FN}}^{\text{te}} = 0$ is equivalent to our hypothesis $\lim_{k \to \infty} FN_k = \emptyset$. □

## B  Statistical guarantees with Sequential Probability Ratio Test

We provide guarantees of the form $P_{\text{A}} \geq \theta_{\text{A}}$ (i.e., the true accuracy value is above $\theta_{\text{A}}$), $P_{\text{FN}} \leq \theta_{\text{FN}}$ and $P_{\text{FP}} \leq \theta_{\text{FP}}$ (i.e., the true rate of FNs and FPs are below $\theta_{\text{FN}}$ and $\theta_{\text{FP}}$, respectively). Being based on hypothesis testing, such guarantees are precise up to arbitrary error bounds $\alpha, \beta \in (0, 1)$, such that the probability of Type-I errors (i.e., of accepting $P_x < \theta_x$ when $P_x \geq \theta_x$, where $x \in \{\text{A}, \text{FN}, \text{FP}\}$) is bounded by $\alpha$, and the probability of Type-II errors (i.e., of accepting $P_x \geq \theta_x$ when $P_x < \theta_x$) is bounded by $\beta$. The pair $(\alpha, \beta)$ is known as the *strength* of the test.

To ensure both error bounds simultaneously, the original test $P_x \geq \theta_x$ vs $P_x < \theta_x$ is relaxed by introducing a small indifference region, i.e., we test the hypothesis $H_0 : P_x \geq p_0$ against $H_1 : P_x \leq p_1$, with $p_0 > p_1$ [24]. Typically, $p_0 = \theta_x + \delta$ and $p_1 = \theta_x - \delta$ for some $\delta > 0$. We use Wald's sequential probability ratio test (SPRT) to provide the above guarantees. SPRT has the important advantage that it does not require a prescribed number of samples to accept one of the two hypotheses, but the decision is made if the available samples provide sufficient evidence. Specifically, after $m$ samples, $H_0$ is accepted if $\frac{p_{1m}}{p_{0m}} \leq B$, while $H_1$ is accepted if $\frac{p_{1m}}{p_{0m}} \geq A$, where $A = (1-\beta)/\alpha$, $B = \beta/(1-\alpha)$ and $\frac{p_{1m}}{p_{0m}} = \frac{p_1^{t_m} \cdot (1-p_1)^{f_m}}{p_0^{t_m} \cdot (1-p_0)^{f_m}}$, where $t_m$ and $f_m$ are, respectively, the numbers of positive and negative samples.

## C  Background on Neural Networks

Let $l$ be the number of layers of an NN, i.e., $l - 1$ hidden layers and one output layer. Let $n_i$ be the number of neurons in layer $i$, $i = 1, \ldots, l$, with $n_0$ being the size of the input vector.

For an input vector $\text{x} \in \mathbb{R}^{n_0}$, the output of the NN classifier is positive if $F(\text{x}) \geq \theta$, negative otherwise, where $F(\text{x})$ is the function represented by the NN and $\theta$ is the classification threshold. Function $F$ is of the following form:

$$F = \text{f}_l \circ \text{f}_{l-1} \circ \ldots \circ \text{f}_1 \circ \text{f}_0,$$

where $\circ$ is the function composition operator, $\text{f}_0$ is the input normalization function, and for $i = 1, \ldots, l$, $\text{f}_i$ is the function computed by the $i$-th layer. The input normalization function typically applies a linear scaling such that the input falls in the range $[-1, 1]$.

The output of layer $i$ results from applying function $\text{f}_i : \mathbb{R}^{n_{i-1}} \to \mathbb{R}^{n_i}$ to the output of the previous layer:

$$\text{f}_i(\text{p}_{i-1}) = \text{g}_i(\text{W}_{i,i-1} \cdot \text{p}_{i-1} + \text{b}_i), \quad i = 1, \ldots, l \tag{5}$$

where $p_{i-1} \in \mathbb{R}^{n_{i-1}}$ is the output vector of layer $i-1$, $W_{i,i-1} \in \mathbb{R}^{n_i \times n_{i-1}}$ is the *weight matrix* that connects $p_{i-1}$ to the neurons of layer $i$, $b_i \in \mathbb{R}^{n_i}$ is the *bias vector* of layer $i$, and $g_i$ is the *activation function* of the neurons of layer $i$.

Weights and biases are the function parameters learned during training, and are typically derived by minimizing the mean square error (or other error functions) between training data and network predictions. The most common optimization algorithm is gradient descent with backpropagation.

We consider two main static configurations of NNs. The first, called **DNN-S**, uses the Tan-Sigmoid activation function tansig for the hidden layers and the Log-Sigmoid activation function logsig for the output layer $l$. Let $z \in \mathbb{R}^{n_i}$ be the argument of the activation function at layer $i$. Then, for neuron $j = 1, \ldots, n_i$, the above activation functions are given by:

$$\text{tansig}(z)_j = \frac{2}{1 + e^{-2 \cdot z_j}} - 1 \quad \text{and} \quad \text{logsig}(z)_j = \frac{1}{1 + e^{-z_j}}. \tag{6}$$

The second configuration, called **DNN-R**, employs the rectified linear unit (ReLU) activation function relu for the hidden layers and the softmax function for the output layer $l$, where

$$\text{relu}(z)_j = \max(0, z_j) \quad \text{and} \quad \text{softmax}(z)_j = \frac{e^{z_j}}{\sum_{k=1}^{n_i} e^{z_k}}. \tag{7}$$

## D  Models and case studies

We briefly introduce the case studies that we used for our evaluation.

*Spiking Neuron.* This model describes the evolution of a neuron's action potential. It is a deterministic HA with two continuous variables, one mode, one transition and nonlinear polynomial dynamics. We consider the unsafe set $U$ defined by $v \leq 68.5$, expressing that the neuron should not undershoot its resting potential. The time bound for the reachability property is $T = 20$.

*Inverted Pendulum.* We consider the classic inverted pendulum on a cart nonlinear system. We consider the unsafe set $U$ defined by $|\theta| > \pi/4$, corresponding to the safety property that keeps the pendulum within $45°$ of the vertical axis. The time bound is $T = 5$.

*Quadcopter Controller.* This model describes the dynamics of a quadcopter [15]. It is a deterministic HA with seven continuous variables, two modes and nonlinear dynamics including trigonometric functions. We consider the safety property that the quadcopter does not crash within time $T = 15$. The set $U$ is defined by the constraint $z \leq 0$, where $z$ is the vehicle's altitude.

*Cruise Control.* It is a nondeterministic HA with 3 continuous variables, 6 modes, 11 transitions, and nonlinear polynomial dynamics. The unsafe set $U$ is defined by $v \leq -1$, which expresses that the vehicle's speed should not be below a reference speed by 1 m/s or more. The reachability time bound is $T = 10$.

*Powertrain Model.* The automotive air-fuel control system model from [19] is a HA with two modes and five nonlinear ODEs describing the dynamics of the engine (throttle air, intake manifold and air-fuel path), coupled with a continuous-time PI controller. $U$ is defined by $|(\lambda - \bar{\lambda})/\bar{\lambda}| > 0.5$, where $\lambda$ is a variable describing the air-fuel ratio and $\bar{\lambda}$ is the target $\lambda$ value. The time bound is $T = 1$. Since the system is already well described in [19], we do not provide the details here.

*Helicopter Controller.* We augment the 28-variable helicopter controller available on SpaceEx website[5] with a variable $z$ denoting the helicopter's altitude. The dynamics of $z$ is given by $\dot{z} = v_z$, where $v_z$ is the vertical velocity and represented by variable $x_8$. The unsafe set $U$ is defined by $z \leq 0$. The time bound is $T = 5$. Since this model is large and publicly available on SpaceEx website, we do not provide the details here.

### D.1 Spiking Neuron

We consider the spiking neuron model on the Flow* website[6]. It is a hybrid system with one mode and one jump. The dynamics is defined by the ODE

$$\begin{cases} \dot{v} &= 0.04v^2 + 5v + 140 - u + I \\ \dot{u} &= a \cdot (b \cdot v - u) \end{cases} \tag{8}$$

The jump condition is $v \geq 30$, and the associated reset is $v' := c \wedge u' := u + d$, where, for any variable $x$, $x'$ denotes the value of $x$ after the reset.

The parameters are $a = 0.02$, $b = 0.2$, $c = -65$, $d = 8$, and $I = 40$ as reported on the Flow* website. We consider the unsafe state set $U = \{(v, u) \mid v \leq 68.5\}$. This corresponds to a safety property that can be understood as the neuron does not undershoot its resting-potential region of $[-68.5, -60]$. The domain for sampling is $68.5 < v \leq 30 \wedge 0 \leq u \leq 25$. The time bound for the reachability property was set to $T = 20$.

### D.2 Inverted Pendulum

We consider the control system for an inverted pendulum on a cart. This is a classic, widely used example of a non-linear system. As shown in Fig. 6, the control input $F$ is a force applied to the cart with the goal of keeping the pendulum in upright position, i.e., $\theta = 0$. The dynamics is given by

$$J \cdot \ddot{\theta} = m \cdot l \cdot g \cdot \sin(\theta) - m \cdot l \cos(\theta) \cdot F \tag{9}$$

where $J$ is the moment of inertia, $m$ is the mass of the pendulum, $l$ is the length of the rod, and $g$ is the gravitational acceleration.

We set $J = 1$, $m = 1/g$, $l = 1$, and let $u = F/g$. Eq. 9 becomes

$$\begin{cases} \dot{\theta} = \omega \\ \dot{\omega} = \sin(\theta) - \cos(\theta) \cdot u \end{cases} \tag{10}$$

---

[5] http://spaceex.imag.fr/
[6] https://flowstar.org/examples/

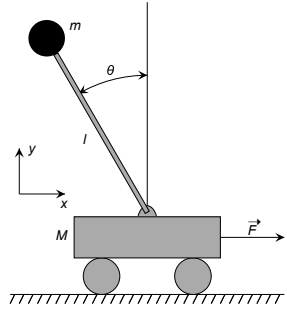

Fig. 6: Schematic of the inverted pendulum on a cart. Source: Wikipedia.

We consider the control law of Eq. 11. Fig. 7 shows an evolution of $\theta$ under this control law. We consider the unsafe state set $U = \{(\theta, \omega) \mid \theta < -\pi/4 \vee \theta > \pi/4\}$. This unsafe region corresponds to the safety property that keeps the pendulum within $45°$ of the vertical axis. The domain for sampling is $\theta \in [-\pi/4, \pi/4] \wedge \omega \in [-1.5, 1.5]$. We used time bound $T = 5$.

$$u = \begin{cases} \dfrac{2 \cdot \omega + \theta + \sin(\theta)}{\cos(\theta)}, & E \in [-1, 1], \mid \omega \mid + \mid \theta \mid \leq 1.85 \\ 0, & E \in [-1, 1], \mid \omega \mid + \mid \theta \mid > 1.85 \\ \dfrac{\omega}{1+ \mid \omega \mid} \cos(\theta), & E < -1 \\ \dfrac{-\omega}{1+ \mid \omega \mid} \cos(\theta), & E > 1 \end{cases} \quad (11)$$

where $E = 0.5 \cdot \omega + (\cos(\theta) - 1)$ is the pendulum energy.

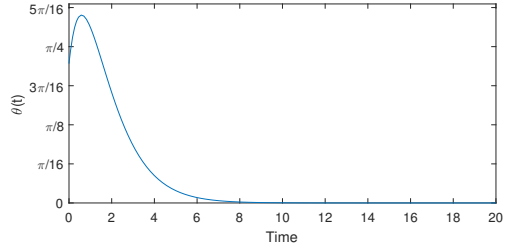

Fig. 7: An evolution of the inverted pendulum state variable $\theta$ from initial state $(\theta_0, \omega_0) = (0.5, 1.0)$.

### D.3 Quadcopter Controller

We consider the quadcopter model of [15]. We consider the safety property that the quadcopter does not crash, i.e., the altitude $z$ is positive. This corresponds to the unsafe state set $U$ defined by $z \leq 0$. This safety property is independent of the state variables $x$, $y$, and $\psi$ (the yaw angle), so we omit them from the model. This hybrid system has two modes that share the following ODEs.

$$\begin{cases} \dfrac{d\omega_x}{dt} = \dfrac{L \cdot k \cdot (\omega_1^2 - \omega_3^2) - (I_{yy} - I_{zz}) \cdot \omega_y \cdot \omega_z}{I_{xx}} \\ \dfrac{d\omega_y}{dt} = \dfrac{L \cdot k \cdot (\omega_2^2 - \omega_4^2) - (I_{zz} - I_{xx}) \cdot \omega_x \cdot \omega_z}{I_{yy}} \\ \dfrac{d\omega_z}{dt} = \dfrac{b \cdot (\omega_1^2 - \omega_2^2 + \omega_3^2 - \omega_4^2) - (I_{xx} - I_{yy}) \cdot \omega_x \cdot \omega_y}{I_{zz}} \\ \dfrac{d\phi}{dt} = \omega_x + \dfrac{\sin(\phi)\sin(\theta)}{\left(\frac{\sin(\phi)^2 \cos(\theta)}{\cos(\phi)} + \cos(\phi)\cos(\theta)\right)\cos(\phi)} \omega_y \\ \qquad + \dfrac{\sin(\theta)}{\frac{\sin(\phi)^2 \cos(\theta)}{\cos(\phi)} + \cos(\phi)\cos(\theta)} \omega_z \\ \dfrac{d\theta}{dt} = -\left(\dfrac{\sin(\phi)^2 \cos(\theta)}{\left(\frac{\sin(\phi)^2 \cos(\theta)}{\cos(\phi)} \omega_y + \cos(\phi)\cos(\theta)\right)\cos(\phi)^2} + \dfrac{1}{\cos(\phi)}\right)\omega_y \\ \qquad - \dfrac{\sin(\phi)\cos(\theta)}{\left(\frac{\sin(\phi)^2 \cos(\theta)}{\cos(\phi)} + \cos(\phi)\cos(\theta)\right)\cos(\phi)} \omega_z \\ \dfrac{dz}{dt} = \dot{z} \end{cases} \quad (12)$$

where the dynamics of $z$ is given by:

$$\text{(mode 1)} \quad \frac{d\dot{z}}{dt} = \frac{g + \cos(\theta) \cdot k \cdot (\omega_1^2 + \omega_2^2 + \omega_3^2 + \omega_4^2) + k \cdot d \cdot \dot{z}}{m} \quad (13)$$

$$\text{(mode 2)} \quad \frac{d\dot{z}}{dt} = \frac{-g - \cos(\theta) \cdot k \cdot (\omega_1^2 + \omega_2^2 + \omega_3^2 + \omega_4^2) - k \cdot d \cdot \dot{z}}{m} \quad (14)$$

The jump from mode 1 to mode 2 happens when $z = 500$, updating variables to $\omega_1' := 0 \wedge \omega_2' := 1 \wedge \omega_3' := 0 \wedge \omega_4' := 1$. The jump from mode 2 to mode 1 occurs at $z = 200$, updating variables to $\omega_1' := 1 \wedge \omega_2' := 0 \wedge \omega_3' := 1 \wedge \omega_4' := 0$.

The parameters are $L = 0.23$, $k = 5.2$, $k \cdot d = 7.5\mathrm{e}{-7}$, $m = 0.65$, $b = 3.13\mathrm{e}{-5}$, $g = 9.8$, $I_{xx} = 0.0075$, $I_{yy} = 0.0075$, $I_{zz} = 0.013$. The domain for sampling is $\omega_x \in [-0.05, 0.05]$, $\omega_y \in [0, 0.1]$, $\omega_z \in [-0.1, 0.1]$, $\phi \in [-0.2, 0.2]$, $\theta \in [-1, 0.4]$, $\dot{z} \in [-150, 150]$, and $z \in [50, 100]$. We chose time bound $T = 15$.

### D.4 Cruise Control

The cruise control is a nondeterministic HA with three continuous variables, six modes, eleven transitions, and nonlinear polynomial dynamics. It is shown in Fig. 8. The continuous variable $v$ denotes the difference between the vehicle's speed and the cruise speed in $m/s$, $x$ is the integral term for the proportional-integral (PI) controller in mode 5, and $t$ is a clock.

In mode 5, the PI controller tries to stabilize $v$ to zero, i.e., to match the vehicle's speed with the cruise speed. Mode 3 and 4 represent the first level of brakes where deceleration increases smoothly from 1.2 to 2.5 $m/s^2$ in mode 4 and stays constant at 2.5 $m/s^2$ in mode 3. Mode 1 and 2 represent the second level of brakes and work in the same way but with higher starting and peak deceleration. Mode 6 constantly accelerates the vehicle. The guards are designed to prevent chattering or Zeno behavior.

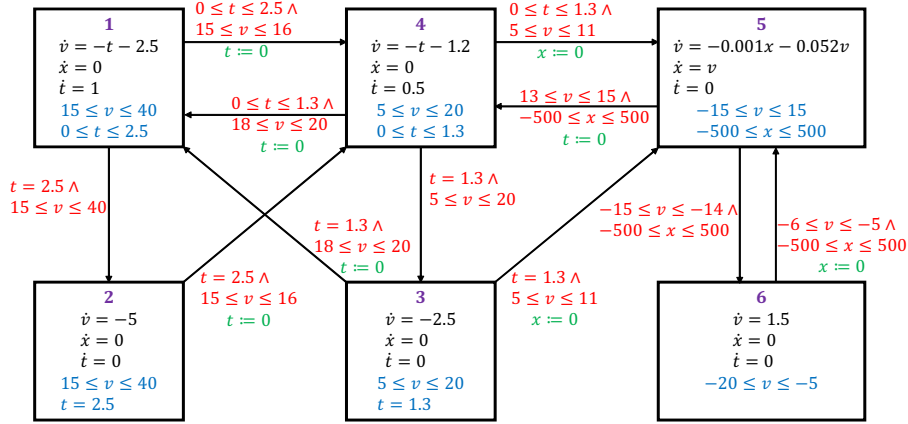

Fig. 8: Hybrid automaton for the cruise control system. Invariants are in blue, guards are in red, and reset mappings are in green.

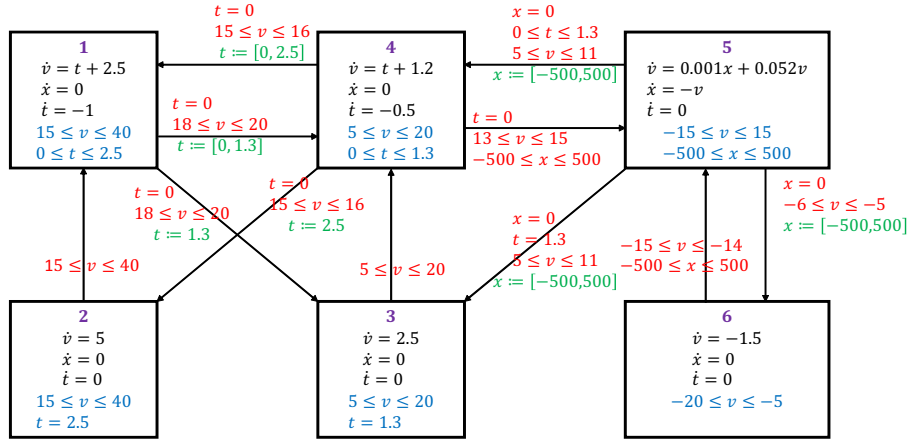

Fig. 9: Reverse hybrid automaton for the cruise control system. Invariants are in blue, guards are in red, and reset mappings are in green.

The unsafe set $U$ is defined by $v \leq -1$, which expresses that the vehicle's speed should not be below a reference speed by 1 m/s or more. The reachability time bound is $T = 10$.

# E  Architecture Comparison

The performance of NNs strongly depends on the chosen architecture. For this reason we evaluated different setups by changing the number of layers and neurons per layer for our DNN-S model. Figure 10 summarizes the accuracy values obtained for each architecture.

We see that as we increase the model complexity there is a tendency to overfit the training data, especially for the quadcopter and cruise case studies. For the neuron and pendulum, we observe that while many model configurations lead to a high accuracy score, having only a single hidden layer and 5 neurons is not sufficient to learn the reachability function. In general, our default configuration of 3 layers and 10 neurons provides a reasonable architectural baseline.

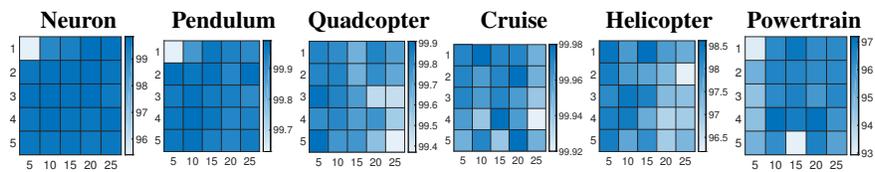

Fig. 10: Heatmaps showing accuracy as a function of number of neurons (horizontal axis) and number of layers (vertical axis).

| **Pendulum (uniform)** | | | | **Pendulum (balanced)** | | |
|---|---|---|---|---|---|---|
| | Acc | FN | FP | Acc | FN | FP |
| **DNN-S** | **99.98** (99.9,100) | **0** (0,0.06) | **0.02** (0,0.1) | **99.89** (99.77,99.96) | **0** (0,0.06) | **0.11** (0.04,0.23) |
| **DNN-R** | 99.93 (99.82,99.98) | 0.04 (0,0.13) | 0.03 (0,0.11) | 99.63 (99.44,99.77) | 0.01 (0,0.08) | 0.36 (0.22,0.55) |
| **SNN** | 99.81 (99.66,99.91) | **0** (0,0.06) | 0.19 (0.09,0.34) | 99.2 (98.94,99.42) | 0.01 (0,0.08) | 0.79 (0.58,1.05) |
| **SVM** | 99.84 (99.7,99.93) | **0** (0,0.06) | 0.16 (0.07,0.3) | 99.37 (99.13,99.56) | **0** (0,0.06) | 0.63 (0.44,0.87) |
| **BDT** | 99.6 (99.4,99.75) | 0.17 (0.08,0.31) | 0.23 (0.12,0.39) | 99.46 (99.24,99.64) | 0.05 (0.01,0.15) | 0.49 (0.32,0.71) |
| **NBOR** | 99.69 (99.51,99.82) | 0.14 (0.06,0.27) | 0.17 (0.08,0.31) | 99.51 (99.29,99.68) | 0.04 (0.00,0.13) | 0.45 (0.29,0.66) |

| **Neuron (uniform)** | | | | **Neuron (balanced)** | | |
|---|---|---|---|---|---|---|
| | Acc | FN | FP | Acc | FN | FP |
| **DNN-S** | **99.81** (99.66,99.92) | 0.1 (0.03,0.22) | **0.09** (0.03,0.2) | **99.83** (99.69,99.92) | **0.12** (0.04,0.25) | **0.05** (0.01,0.15) |
| **DNN-R** | 99.52 (99.31,99.68) | 0.29 (0.17,0.46) | 0.18 (0.08,0.33) | 99.48 (99.26,99.65) | 0.24 (0.13,0.4) | 0.28 (0.16,0.45) |
| **SNN** | 99.17 (98.9,99.39) | 0.43 (0.28,0.63) | 0.4 (0.25,0.6) | 98.89 (98.59,99.15) | 0.69 (0.49,0.94) | 0.42 (0.27,0.62) |
| **SVM** | 98.73 (98.41,99.01) | 0.75 (0.54,1.01) | 0.52 (0.35,0.74) | 98.63 (98.3,98.92) | 0.78 (0.57,1.04) | 0.59 (0.41,0.82) |
| **BDT** | 99.3 (99.05,99.5) | 0.37 (0.23,0.56) | 0.33 (0.2,0.51) | 99.07 (98.79,99.3) | 0.45 (0.29,0.66) | 0.48 (0.32,0.69) |
| **NBOR** | 97.03 (96.56,97.45) | 1.75 (1.43,2.12) | 1.22 (0.95,1.54) | 96.95 (96.47,97.38) | 1.62 (1.31,1.98) | 1.43 (1.14,1.77) |

| **Quadcopter (uniform)** | | | | **Quadcopter (balanced)** | | |
|---|---|---|---|---|---|---|
| | Acc | FN | FP | Acc | FN | FP |
| **DNN-S** | 99.83 (99.69,99.92) | 0.1 (0.03,0.22) | 0.07 (0.02,0.18) | **99.82** (99.67,99.92) | 0.04 (0,0.13) | **0.14** (0.06,0.27) |
| **DNN-R** | **99.89** (99.77,99.96) | **0.06** (0.01,0.16) | **0.05** (0.01,0.15) | 99.67 (99.49,99.8) | 0.09 (0.03,0.2) | 0.24 (0.13,0.4) |
| **SNN** | 99.85 (99.71,99.94) | 0.08 (0.02,0.19) | 0.07 (0.02,0.18) | 99.49 (99.27,99.66) | **0.01** (0,0.08) | 0.5 (0.33,0.72) |
| **SVM** | 97.33 (96.88,97.73) | 0.69 (0.49,0.94) | 1.98 (1.63,2.37) | 96.93 (96.45,97.36) | 0.2 (0.1,0.35) | 2.87 (2.45,3.33) |
| **BDT** | 99.52 (99.31,99.68) | 0.2 (0.1,0.35) | 0.28 (0.16,0.45) | 99.36 (99.12,99.55) | 0.22 (0.11,0.38) | 0.42 (0.27,0.62) |
| **NBOR** | 99.53 (99.32,99.69) | 0.25 (0.14,0.41) | 0.22 (0.11,0.38) | 99.11 (98.83,99.34) | 0.56 (0.38,0.79) | 0.33 (0.2,0.51) |

| **Cruise (uniform)** | | | | **Cruise (balanced)** | | |
|---|---|---|---|---|---|---|
| | Acc | FN | FP | Acc | FN | FP |
| **DNN-S** | 99.95 (99.85,99.99) | 0.01 (0,0.08) | 0.04 (0,0.13) | 99.94 (99.84,99.99) | **0** (0,0.06) | 0.06 (0.01,0.16) |
| **DNN-R** | **99.98** (99.9,100) | **0** (0,0.06) | 0.02 (0,0.1) | **99.95** (99.85,99.99) | **0** (0,0.06) | **0.05** (0.01,0.15) |
| **SNN** | 99.84 (99.7,99.93) | 0.15 (0.06,0.29) | **0.01** (0,0.08) | 99.6 (99.4,99.75) | **0** (0,0.06) | 0.4 (0.25,0.6) |
| **SVM** | 99.88 (99.75,99.96) | 0.1 (0.03,0.22) | 0.02 (0,0.1) | 99.61 (99.41,99.76) | **0** (0,0.06) | 0.39 (0.24,0.59) |
| **BDT** | 99.84 (99.7,99.93) | 0.08 (0.02,0.19) | 0.08 (0.02,0.19) | 99.9 (99.78,99.97) | 0.03 (0,0.11) | 0.07 (0.02,0.18) |
| **NBOR** | 99.49 (99.27,99.66) | 0.33 (0.2,0.51) | 0.18 (0.08,0.33) | 99.47 (99.25,99.64) | 0.11 (0.04,0.23) | 0.42 (0.27,0.62) |

Table 5: Accuracy (Acc) and FN rate of the state classifiers on the pendulum, neuron, quadcopter and cruise case studies. All results are expressed as percentages and are reported as $a$ $(b, c)$, where $a$ is the sample mean and $(b, c)$ is the (open) 99% confidence interval (conservative over-approximation to the closest decimal). For each measure and sampling method, the best result is highlighted in bold.

| | Powertrain (uniform) | | | | Powertrain (balanced) | | |
|---|---|---|---|---|---|---|---|
| | Acc | FN | FP | | Acc | FN | FP |
| **DNN-S** | **96.68** (96.19,97.13) | 1.28 (1,1.6) | **2.04 (1.69,2.44)** | **DNN-S** | **97.2** (96.74,97.61) | **0.86 (0.64,1.13)** | **1.94 (1.6,2.33)** |
| **DNN-R** | 96.21 (95.69,96.69) | **1.08 (0.83,1.38)** | 2.71 (2.3,3.16) | **DNN-R** | 96.07 (95.54,96.56) | 1.24 (0.97,1.56) | 2.69 (2.29,3.14) |
| **SNN** | 96.02 (95.48,96.51) | 1.37 (1.08,1.7) | 2.61 (2.21,3.05) | **SNN** | 95.21 (94.63,95.75) | 1.79 (1.46,2.17) | 3 (2.57,3.47) |
| **SVM** | 92.26 (91.54,92.94) | 3.48 (3.02,3.98) | 4.26 (3.75,4.81) | **SVM** | 91.84 (91.19,92.54) | 3.3 (2.85,3.79) | 4.86 (4.32,5.45) |
| **BDT** | 95.59 (95.03,96.11) | 2.19 (1.83,2.6) | 2.22 (1.85,2.63) | **BDT** | 95.86 (95.31,96.36) | 2.4 (2.02,2.83) | 1.74 (1.42,2.11) |
| **NBOR** | 71.44 (70.26,72.6) | 14.51 (13.61,15.44) | 14.05 (13.16,14.97) | **NBOR** | 71.33 (70.15,72.49) | 13.99 (13.1,14.91) | 14.68 (13.78,15.62) |

| | Helicopter (uniform) | | | | Helicopter (balanced) | | |
|---|---|---|---|---|---|---|---|
| | Acc | FN | FP | | Acc | FN | FP |
| **DNN-S** | **98.49 (98.14,98.79)** | **0.84 (0.62,1.11)** | **0.67 (0.47,0.92)** | **DNN-S** | **98.24 (97.87,98.57)** | **0.79 (0.58,1.05)** | 0.97 (0.73,1.26) |
| **DNN-R** | 98 (97.61,98.35) | 0.96 (0.72,1.25) | 1.04 (0.79,1.34) | **DNN-R** | 97.91 (97.51,98.27) | 1.2 (0.93,1.51) | **0.89 (0.66,1.17)** |
| **SNN** | 97.69 (97.27,98.06) | 1.25 (0.98,1.57) | 1.06 (0.81,1.36) | **SNN** | 97.58 (97.15,97.96) | 1.16 (0.9,1.47) | 1.26 (0.99,1.58) |
| **SVM** | 95.58 (95.02,96.1) | 2.42 (2.04,2.85) | 2 (1.65,2.39) | **SVM** | 95.36 (94.79,95.89) | 1.85 (1.52,2.23) | 2.79 (2.38,3.25) |
| **BDT** | 80.07 (79.02,81.09) | 9.8 (9.04,10.6) | 10.13 (9.36,10.94) | **BDT** | 79.03 (77.96,80.07) | 10.26 (9.49,11.07) | 10.71 (9.92,11.54) |
| **NBOR** | 67.39 (66.16,68.6) | 16.98 (16.02,17.97) | 15.63 (14.7,16.59) | **NBOR** | 65.18 (63.94,66.41) | 17.48 (16.51,18.48) | 17.34 (16.37,18.34) |

Table 6. Accuracy (Acc) and FN rate of the state classifiers on the powertrain and helicopter case studies. See caption of table 6 for details on notation.